\documentclass[conference]{IEEEtran}
\IEEEoverridecommandlockouts
\usepackage{cite}
\usepackage{amsmath,amssymb,amsfonts}
\usepackage{graphicx}
\usepackage[caption=false]{subfig}
\usepackage[utf8]{inputenc}
\usepackage[T1]{fontenc}
\usepackage{bbm}

\usepackage{booktabs}
\usepackage{xcolor}
\usepackage{tikz}
\usepackage{float}
\usepackage[noend]{algpseudocode}
\usepackage[warn]{textcomp}
\usepackage{tikz}

\usepackage{siunitx}
\usepackage{scalerel}
\usepackage{balance}
\usepackage{tabularx}
\usepackage{multirow}
\usepackage{algorithm}
\usepackage{comment}
\usepackage{hyperref} 
\usetikzlibrary{svg.path}
\usepackage{booktabs, makecell}
\newcommand\Algphase[1]{
\vspace*{-.7\baselineskip}\Statex\hspace*{\dimexpr-\algorithmicindent-2pt\relax}\rule{\columnwidth}{0.4pt}
\Statex\hspace*{-\algorithmicindent}\textbf{#1}
\vspace*{-.7\baselineskip}\Statex\hspace*{\dimexpr-\algorithmicindent-2pt\relax}\rule{\columnwidth}{0.4pt}
}
\definecolor{orcidlogocol}{HTML}{A6CE39}
\tikzset{
  orcidlogo/.pic={
    \fill[orcidlogocol] svg{M256,128c0,70.7-57.3,128-128,128C57.3,256,0,198.7,0,128C0,57.3,57.3,0,128,0C198.7,0,256,57.3,256,128z};
    \fill[white] svg{M86.3,186.2H70.9V79.1h15.4v48.4V186.2z} svg{M108.9,79.1h41.6c39.6,0,57,28.3,57,53.6c0,27.5-21.5,53.6-56.8,53.6h-41.8V79.1z M124.3,172.4h24.5c34.9,0,42.9-26.5,42.9-39.7c0-21.5-13.7-39.7-43.7-39.7h-23.7V172.4z} svg{M88.7,56.8c0,5.5-4.5,10.1-10.1,10.1c-5.6,0-10.1-4.6-10.1-10.1c0-5.6,4.5-10.1,10.1-10.1C84.2,46.7,88.7,51.3,88.7,56.8z};
  }
}
\newcommand\orcidicon[1]{\href{https://orcid.org/#1}{\mbox{\scalerel*{
\begin{tikzpicture}[yscale=-1,transform shape]
\pic{orcidlogo};
\end{tikzpicture}
}{|}}}}
\newcommand\copyrighttext{
\footnotesize \textcopyright 2021 IEEE. Personal use of this material is permitted. Permission from IEEE must be obtained for all other uses, in any current or future media, including reprinting/republishing this material for advertising or promotional purposes, creating new collective works, for resale or redistribution to servers or lists, or reuse of any copyrighted component of this work in other works.}
\newcommand\copyrightnotice{\begin{tikzpicture}[remember picture,overlay]
\node[anchor=south,yshift=10pt] at (current page.south) {\fbox{\parbox{\dimexpr\textwidth-\fboxsep-\fboxrule\relax}{\copyrighttext}}};
\end{tikzpicture}}

\begin{document}
\bstctlcite{IEEEexample:BSTcontrol}
\title{\LARGE \bf Remote Electrical Tilt Optimization via Safe Reinforcement Learning}
\author{\IEEEauthorblockN{
Filippo Vannella\IEEEauthorrefmark{1}\IEEEauthorrefmark{2}\textsuperscript{\textsection},\orcidicon{0000-0002-7668-0650},
Grigorios Iakovidis\IEEEauthorrefmark{1}\IEEEauthorrefmark{2}\textsuperscript{\textsection},\orcidicon{0000-0001-7168-3169}, 
Ezeddin Al Hakim\IEEEauthorrefmark{2},\orcidicon{0000-0002-6207-326X},
Erik Aumayr\IEEEauthorrefmark{3},\orcidicon{0000-0002-7071-8929},
Saman Feghhi\IEEEauthorrefmark{3}\orcidicon{0000-0003-1988-9615} 
\\
\IEEEauthorrefmark{1}KTH Royal Institute of Technology, Stockholm, Sweden\\
\IEEEauthorrefmark{2}Ericsson Research, Stockholm, Sweden\\
\IEEEauthorrefmark{3}Network Management Research Lab, LM Ericsson, Athlone, Ireland\\
Email: \{vannella, griiak\}@kth.se, \{ezeddin.al.hakim, erik.aumayr, saman.feghhi\}@ericsson.com}}
\maketitle
\copyrightnotice
\begingroup\renewcommand\thefootnote{\textsection}
\footnotetext{Equal contribution.}
\thispagestyle{empty}
\pagestyle{empty}

\begin{abstract}
Remote Electrical Tilt (RET) optimization is an efficient method for adjusting the vertical tilt angle of Base Stations (BSs) antennas in order to optimize Key Performance Indicators (KPIs) of the network. Reinforcement Learning (RL) provides a powerful framework for RET optimization because of its self-learning capabilities and adaptivity to environmental changes. However, an RL agent may execute unsafe actions during the course of its interaction, i.e., actions resulting in undesired network performance degradation. Since the reliability of services is critical for Mobile Network Operators (MNOs), the prospect of performance degradation has prohibited the real-world deployment of RL methods for RET optimization. In this work, we model the RET optimization problem in the Safe Reinforcement Learning (SRL) framework with the goal of learning a tilt control strategy providing performance improvement guarantees with respect to a safe baseline. We leverage a recent SRL method, namely Safe Policy Improvement through Baseline Bootstrapping (SPIBB), to learn an improved policy from an offline dataset of interactions collected by the safe baseline. Our experiments show that the proposed approach is able to learn a safe and improved tilt update policy, providing a higher degree of reliability and potential for real-world network~deployment.
\end{abstract}

\section{Introduction}
\label{sec:intro}
Network performance optimization represents one of the major challenges for Mobile Network Operators (MNOs). Due to the growing complexity of modern networks, MNOs increasingly face the need to satisfy consumer demand that is highly variable in both the spatial and temporal domains. In order to provide a high level of Quality of Service (QoS) efficiently to each User Equipment (UE), networks must adjust their configuration in an automatic and timely manner. 
The antenna \textit{downtilt}, defined as the inclination angle of the antenna's radiating beam with respect to the horizontal plane, plays a crucial role in the network performance optimization. Remote Electrical Tilt (RET) optimization is a powerful technique to remotely control the downtilt of Base Stations (BSs) antennas in order to optimize Key Performance Indicators (KPIs) while efficiently utilizing network resources.

In this work, we focus on RET optimization techniques using Reinforcement Learning (RL) methods \cite{sutton2018reinforcement}. RL is a decision-making paradigm used to solve complex problems by direct interaction with an environment. RL offers an effective framework for RET optimization due to its self-learning capabilities and adaptivity to environmental changes. Because of these potentialities, during the past decade, there has been a significant amount of research into the field of RET optimization using RL methods \cite{Balevi2019OnlineLearning, Guo2013Spectral-andLearning, Razavi10, Fan2014Self-optimizationLearning}. 
The authors of \cite{Balevi2019OnlineLearning} tackle RET optimization using a $Q$-learning method with linear function approximation, while   \cite{Guo2013Spectral-andLearning} employ an RL technique based on Boltzmann exploration to maximize the throughput fairness while increasing energy efficiency. Another class of methods combines RL and Fuzzy Logic in the Fuzzy RL (FRL) framework, using a Fuzzy inference system to discretize the states and model the intrinsic uncertainty in cellular networks \cite{Razavi10,Fan2014Self-optimizationLearning}.
Although RL methods have shown great promise by outperforming traditional control-based \cite{Tall15} or Rule-Based (RB) \cite{Buenostado17} approaches, such methods utilize traditional exploration-exploitation strategies that do not preclude the possibility of selecting unsafe actions, failing to provide any performance guarantee. 


Most RL methods ignore their inherent \textit{unsafe} nature. During the RL interaction, the agent is faced with a fundamental dilemma: it must decide whether to prioritize further gathering of information (\textit{exploration}) or to execute the best action given the current knowledge (\textit{exploitation}). The primary source of unsafety in RL methods can be imputed to the agent's exploration, resulting in undesired network performance degradation. Since the reliability of services is a critical feature for MNOs, the prospect of performance degradation has in fact prohibited the real-world deployment of RL methods for RET optimization, limiting their use to simulated network environments. For this reason, \textit{safety} has become an essential feature in RET optimization approaches using RL techniques, since it represents the primary obstacle for the deployment of such techniques, and precludes the possibility to achieve improved network performance. 

Safe Reinforcement Learning (SRL) aims at solving RL problems in which a minimum performance level must be guaranteed during learning and deployment. Recently, there have been many attempts in the field of SRL to design safe RL algorithms. We refer the interested reader to \cite{Garcia2015} for a complete survey on SRL. In particular, we focus on SRL methods falling in the domain of \textit{Offline RL} \cite{Lange2012BatchLearning, Thomas2015HighImprovement, Laroche2019SafeBootstrapping}, in which the goal is to devise an optimal policy from offline data, without further interaction with the real environment. The dataset is collected by a \textit{baseline policy}, considered to be safe and sub-optimal. There exist a limited amount of work dealing with SRL for RET optimization. The authors of \cite{Vannella20} apply an off-policy Contextual (CB) formulation to learn a RET policy completely offline from real-world network data. \newpage \noindent However, the CB formulation imposes severe approximations to the RET optimization model, and the off-policy learning performance is highly susceptible to the data quality. In \cite{Vairamuth20}, the author consider an SRL method based on the use of an actor-critic model executing actions for which the probability is $\varepsilon$-close to the actions of a safe baseline policy. This method may be excessively conservative if the safety threshold $\varepsilon$ is not correctly tuned, and it does not have convergence guarantees.

In this work, we formulate the RET optimization problem in the SRL framework and apply an SRL method to devise a safe downtilt update policy. In particular, we leverage a recent Offline RL method, namely Safe Policy Improvement with Baseline Bootstrapping (SPIBB) \cite{Laroche2019SafeBootstrapping}, in order to learn a policy having performance improvement guarantees over a baseline policy. We empirically evaluate the SRL approach to RET optimization for different types of baseline policies and over different worst-case scenarios. The outcomes of our experiments show that the proposed SRL approach is able to learn a safe and improved downtilt update policy, providing a higher degree of reliability and potential for real-world network deployment. 


\section{Background}
\label{sec:background}
In this section, we introduce the network environment model considered for RET optimization, the background on the RL and SRL framework, and algorithms in both frameworks utilized in the remainder of the paper.
\begin{figure}[!htb]
    \centering
    \includegraphics[scale=0.27]{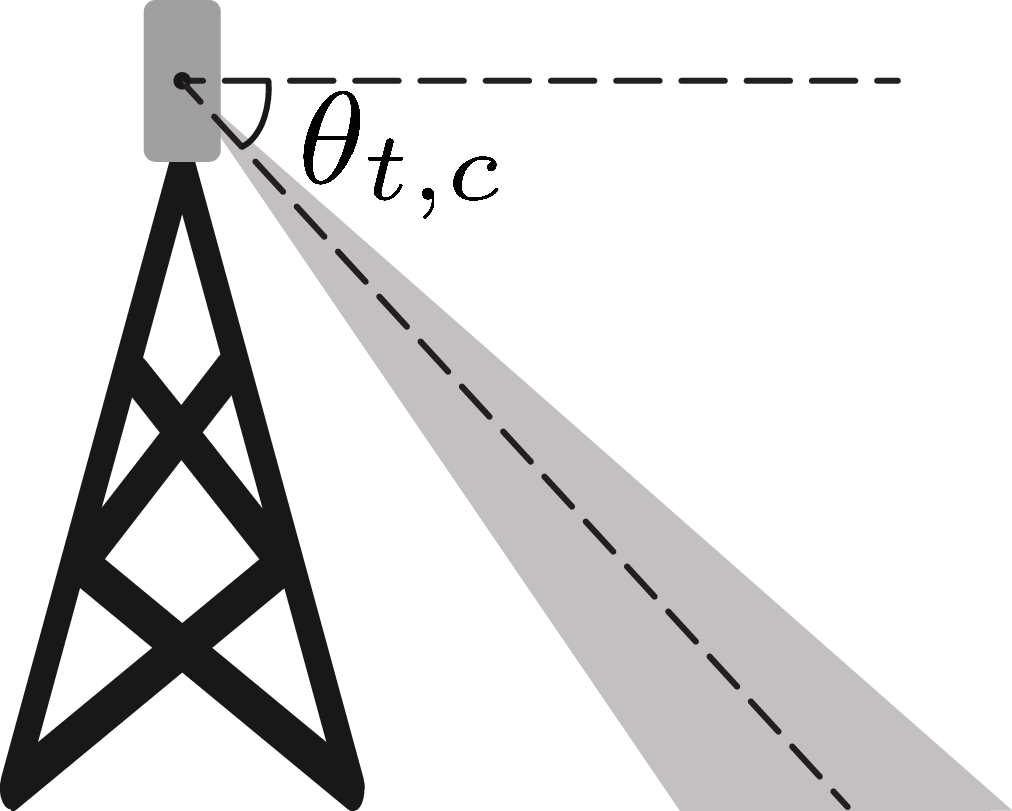}
    \caption{Representation of the downtilt $\theta_{t,c}$ at time $t$ and for cell $c$.}
    \label{fig:downtilt}
\end{figure}

\subsection{Network environment model}
\label{sec:model_env}
We consider a geographical area covered by a sectorized mobile network environment containing $\mathcal{B}$ BSs and $\mathcal{C}$ cells. The network environment is populated by $\mathcal{U}$ UE, uniformly distributed in the considered area. Time is indexed by $t = 1,2,\dots$, and $\theta_{t,c}$, represented in  Fig. \ref{fig:downtilt}, denotes the \textit{downtilt} of the antenna at time $t$ and for cell $c$. The RET problem formulation used in this paper follows previous works in the literature based on Coverage-Capacity Optimization (CCO) \cite{Buenostado17, Vannella20}. The goal in CCO is to maximize both network \textit{coverage} and \textit{capacity}, while minimizing inter-cell \textit{interference}. We model such interference term through the \textit{quality} KPI.

The \textit{coverage} KPI measures to which extent the area of interest is adequately covered by the network and is computed using measurements of the Reference Signal Received Power (RSRP) and the Radio Resource Control (RRC) failure rate. The \textit{capacity} KPI measures the amount of UE that the cell can handle simultaneously (e.g., in terms of number of calls) with a reference bit rate. We consider the network measurements for capacity and coverage KPIs to be collected at time $t$ and for cell $c$, and we denote them by $\textsc{Cov}_{t,c}$ and $\textsc{Cap}_{t,c}$ respectively. The downtilt determines a trade-off between coverage and capacity: low downtilt values result in a larger covered area and smaller capacity due to lower signal power received; in contrast, high downtilt values result in a smaller covered area with increased capacity, but it risks to create \textit{coverage holes} in the cells. The joint optimization of these mutually opposed KPIs ensures that the targeted area remains covered while maximizing the capacity. It is critical to note that the cell's performance is strongly influenced by the downtilt of neighboring cells, that introduce coupled interference between cells. In our formulation, the effect of negative interference from neighboring cells is modeled by the \textit{quality} KPI, denoted as $\textsc{Qual}_{t,c}$ at time $t$ and for cell $c$. Quality is calculated using the \textit{cell overshooting} and \textit{cell overlapping} indicators defined in \cite{Buenostado17}, which in turn
depends on measurements of the RSRP level differences between a cell $c$ and its neighbors. 

In this paper, we consider the class of RET control strategies whose input consists of the \textit{downtilt}, \textit{coverage}, \textit{capacity}, and \textit{quality} KPIs for cell $c$ and time step $t$, and whose output consists of one between the actions \textit{increase the downtilt}, \textit{decrease the downtilt}, or \textit{keep the same downtilt}. The RET control strategy manages the antenna tilt on a cell-by-cell basis and is global, i.e., the same control strategy is executed independently for all cells $c = 1,\dots, \mathcal{C}$.

\subsection{The RL framework}
\label{sec:rl_model}
\renewcommand{\thefootnote}{\textreferencemark} 
RL \cite{sutton2018reinforcement} is a powerful algorithmic framework in which an agent learns an optimal control strategy by direct interaction with an unknown environment. The RL agent iteratively observes a representation of the state of the environment, executes an action, receives a scalar reward, and transitions to a new state. The goal of the agent is to learn a policy maximizing the expected cumulative reward.

Formally, an RL problem is defined by a Markov Decision Process (MDP), described by a tuple $\langle \mathcal{S}, \mathcal{A}, R, P, \gamma,p_0\rangle$, where $\mathcal{S}$ is the state space, $\mathcal{A}$ is the action space, $R(\cdot|s,a)$ is the reward probability distribution, $P(\cdot|s,a)$ is the transition probability distribution, $\gamma$ is the discount factor, and $p_0$ is the initial state distribution. We assume the state space is continuous $\mathcal{S}\subseteq\mathbb{R}^d$ and the action space $\mathcal{A}$ is discrete.

The policy of the agent $\pi(\cdot|s): \mathcal{S} \to \Delta_\mathcal{A}$ is a mapping from states to probability distributions over actions\footnote{This formulation includes the case of a deterministic policy $\pi:\mathcal{S}\to \mathcal{A}$.}, defining the agent's control strategy. The RL interaction proceeds as follows: at each time instant $t = 1,2,\dots$, the agent receives a representation of the environment state $s_t \in \mathcal{S}$, selects an action $a_t \sim \pi(\cdot|s_t) \in \mathcal{A}$, receives a stochastic reward ${r_t\sim R\left(\cdot| s_t,a_t\right)}$, and transitions to the next state ${s'_{t}\sim P\left(\cdot\middle|s_t,a_t\right)}$. In the RL setting, the goal of the agent is to devise an optimal policy, i.e., a policy maximizing the expected cumulative reward over a predefined period of time.

\subsection{$Q$-learning and Deep $Q$-Network}
$Q$-learning \cite{Watkins89} is one of the most popular RL algorithms. It aims at estimating the state-action value function (a.k.a. $Q$-function) when following policy $\pi$, defined as 
\begin{equation*}
Q\left(s,a\right)=\mathbb{E}_\pi\left[\sum_{\tau=0}^{\infty}{\gamma^\tau r_{t+\tau+1}\Big| s_t=s,a_t=a}\right],
\end{equation*}
where the expected value operator $\mathbb{E}_\pi[\cdot]$ is defined with respect to actions sampled by policy $\pi$. In its tabular form, the $Q$-function is stored as matrix whose entries $(s,a)\in\mathcal{S}\times \mathcal{A}$ correspond to the $Q$-value of a given state-action pair. At each time step $t$, 
the state-action value is updated as
\begin{equation*}
Q\left(s_{t}, a_{t}\right) \leftarrow Q\left(s_{t}, a_{t}\right)+\eta\left[r_t
+\gamma \max_{a^{\prime}} Q\left(s_t^{\prime}, a^\prime\right)-Q\left(s_{t}, a_{t}\right)\right],
\end{equation*}
with learning rate $\eta \in (0,1]$.
In the case of continuous state space, the $Q$-function is parametrized through a function approximator $Q_w(s,a)$, where $w \in \mathbb{R}^\Omega$ is the parameter vector, allowing to generalize the $Q$-function representation to unseen states. When an Artificial Neural Network (ANN) parametrization is chosen, the $Q$-learning algorithm is also known as Deep~$Q$-Network~(DQN)~\cite{Mnih15}. DQN usually makes use of \textit{experience replay}, a biologically inspired learning technique that consists in storing trajectories of experience in a dataset and reusing them in future training steps in order to remove the correlation between the samples. The policy used to collect the data, is usually designed as an exploratory policy (e.g. $\varepsilon$-greedy policy) in order to ensure sufficient exploration of the state-action space. 
DQN uses optimization methods (e.g. Stochastic Gradient Descent (SGD)) to minimize the Mean Squared Error (MSE) between the target ${y_t = r_t + \gamma \max_{a'}Q_w(s'_{t},a')}$ and the parametrized $Q$-function $Q_w(s,a)$. The weight update step is executed as 
\begin{equation*}
    w \leftarrow w - \eta \cdot\nabla_w \left(y_t-Q_w\left(s_{t}, a_{t}\right)\right)^{2},
\end{equation*}
where $\nabla_w$ denotes the gradient operator with respect to the weight vector $w$.



\subsection{SPIBB-DQN}
\label{sec:safe_rl_baseline}
In the context of RL, an optimal policy is usually learned in a \textit{trial-and-error} fashion by direct interaction with the environment. In the course of such interaction, the agent will explore sub-optimal regions of the state-action space. In some domains, sub-optimal exploration may result in unacceptable risky circumstances or performance degradation. DQN-based algorithms have shown superior performance in many applications. However, due to the inherent exploratory nature of $Q$-learning policies, DQN fails to provide safety guarantees. 

One of the most effective ways to achieve safety in RL is to rely on a \textit{safe baseline policy} $\pi_b$ and on the \textit{baseline dataset} $\mathcal{D}_{\pi_b} = \{(s_n,a_n,r_n,s'_{n})\}_{n=1}^N$ of previous interactions of $\pi_b$ with the environment. The goal is to learn a policy $\pi$ having better performance (e.g. in terms of $Q$-function) with respect to the baseline policy $\pi_b$ in an offline manner from $\mathcal{D}_{\pi_b}$, and without further interaction with the environment.

We make use of the SPIBB method, as presented in \cite{Laroche2019SafeBootstrapping}. The high-level idea of SPIBB is to rely on the guidance provided by a safe baseline policy $\pi_b$ for state-action pairs $(s,a)$ that are rarely observed in the baseline dataset $\mathcal{D}_{\pi_b}$. Specifically, the method first constructs a \textit{bootstrapped set} containing a set of state action pairs ${(s,a) \in \mathcal{S} \times \mathcal{A}}$ that have been observed with a low frequency in $\mathcal{D}_{\pi_b}$. Formally, let us denote by $N_{\mathcal{D}_{\pi_b}}(s,a) = \sum_{n = 1}^N \mathbbm{1}\{s_n = s,a_n = a\}$ the \textit{state-action count} of dataset $\mathcal{D}_{\pi_b}$, and define the \textit{bootstrapped set} as
\begin{equation}
\label{eq:bootstrapped}
    \mathfrak{B}_{\pi_b} = \{(s,a)\in \mathcal{S} \times \mathcal{A}:N_{\mathcal{D}_{\pi_b}}(s,a) < N_\wedge \},
\end{equation}
where $N_\wedge$ is a hyperparameter controlling the level of safety. In particular we focus on the SPIBB-DQN algorithm, which can be considered as a safe variation of the DQN through baseline bootstrapping. The main variation with respect to DQN is in the target of the $Q$-function update, executed on a batch of samples $\{(s_{\beta},a_{\beta},r_{\beta},s'_{\beta})\}_{\beta = 1}^{N_B}$ and for update index~$j$~as  
\begin{equation}
\begin{aligned}
y^{(j)}_{\beta}& =  \underbrace{r_{\beta}}_{(i)}+\underbrace{\gamma \sum_{a^{\prime} \mid\left(s_{\beta}^{\prime}, a^{\prime}\right) \in \mathfrak{B}_{\pi_b}} \pi_{b}\left(a^{\prime} \mid s_{\beta}^{\prime}\right) Q^{(j)}\left(s_{\beta}^{\prime}, a^{\prime}\right)}_{(ii)} \\&
+\underbrace{\gamma\sum_{a^{\prime} \mid\left(s_{\beta}^{\prime}, a^{\prime}\right) \notin \mathfrak{B}_{\pi_b}} \pi_{b}\left(a^{\prime} \mid s_{\beta}^{\prime}\right)\max_{a^{\prime} \mid\left(s_{\beta}^{\prime}, a^{\prime}\right) \notin \mathfrak{B}_{\pi_b}} Q^{(j)}\left(s_{\beta}^{\prime}, a^{\prime}\right)}_{(iii)}.
\label{eq:SPIBB_update}
\end{aligned}
\end{equation}
This update consists of three elements: $(i)$ the immediate return observed in the dataset's recorded transition, $(ii)$ the estimated return of the bootstrapped actions (where the agent must follow the policy of the baseline), and $(iii)$ the maximum over the return estimates of the non-bootstrapped actions, where the agent is free to exploit the optimal strategy.

It is easy to observe that for $N_\wedge \to \infty$, the bootstrapped set contains the whole state-action space, and the training policy converges to the baseline policy. Meanwhile, for $N_\wedge \to 0$, the bootstrapped set is empty, and the SPIBB-DQN degenerates into a vanilla DQN with no safety-preserving influence from the baseline. Thus, the safety parameter $N_\wedge$ must be chosen large enough to ensure safety, and small enough to ensure improvement.

\section{RET Optimization via SRL}
\label{sec:RET_safe}
In this section, we formulate the RET optimization problem as an SRL problem, and we illustrate how to apply the SPIBB-DQN algorithm to RET optimization.

\begin{figure}[ht!]
\begin{centering}
    \includegraphics[width =\columnwidth]{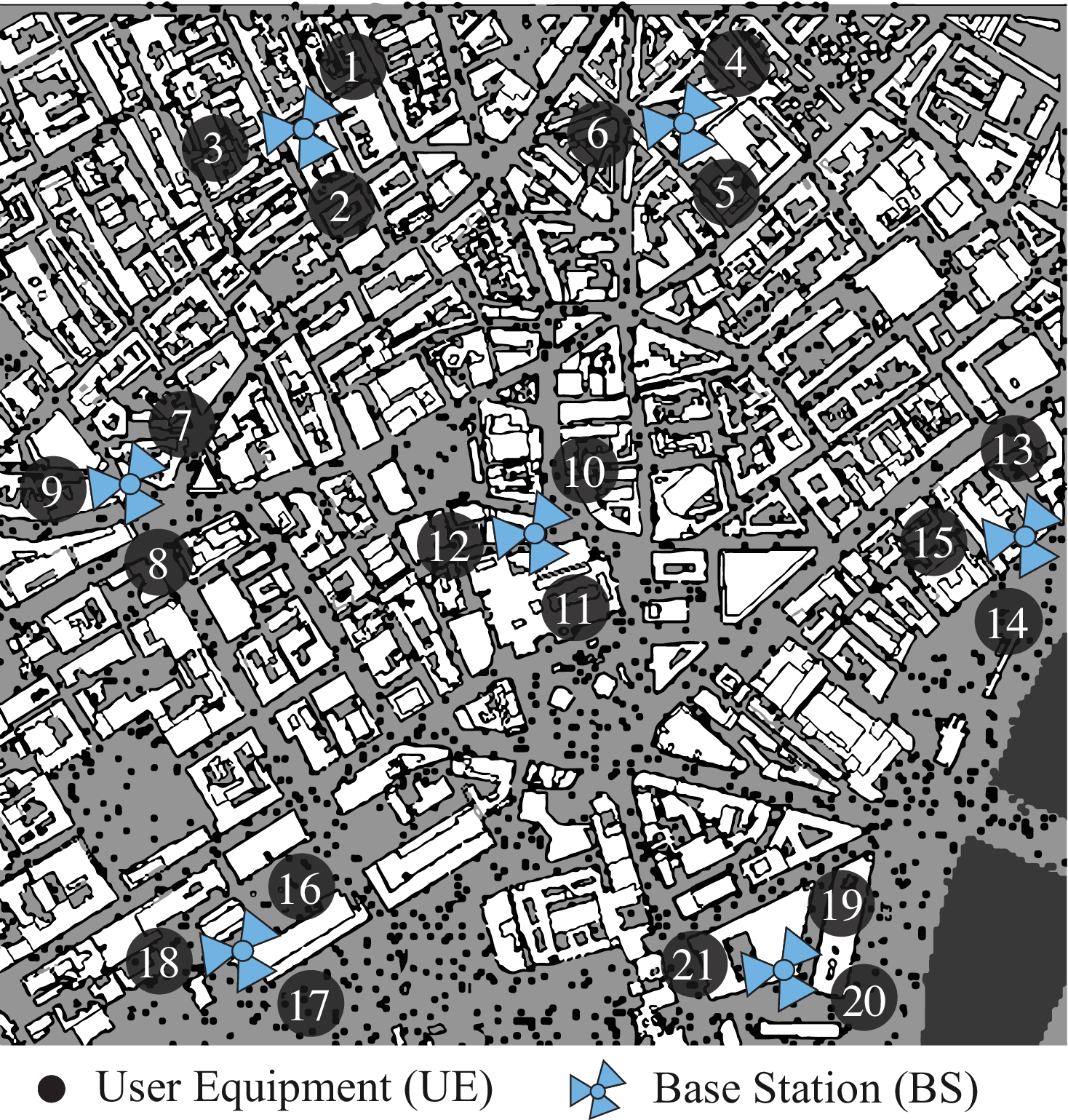}
    \caption{Map of the simulated mobile network environment.}
    \label{fig:network_env}
\end{centering}
\end{figure}
\subsection{SRL formulation for RET optimization}
\label{subsec:SRL_form}
We formulate the RET optimization as an SRL problem by specifying its constituent elements:
\begin{itemize}
  \item \textit{Environment} $\mathcal{E}$: consists of a simulated mobile network in an urban environment, as described in Section \ref{sec:model_env}, and represented in Fig. \ref{fig:network_env}. More information about the environment parameters are reported in Table \ref{tab:simulator_setup}. 
  \item \textit{State space} $\mathcal{S} \subseteq [0,1]^4$: comprises normalized values of \textit{downtilt}, \textit{coverage}, \textit{capacity} and \textit{quality} for a given cell $c$., i.e. $s_{t,c} = [\Bar{\theta}_{t,c},\textsc{Cov}_{t,c},\textsc{Cap}_{t,c}, \textsc{Qual}_{t,c}]$. 
  \item \textit{Action space} $\mathcal{A} =\{a_{-\delta},a_{0},a_{\delta}\}$: consists of three discrete actions: down-tilt, no-change, or up-tilt the current downtilt of magnitude $\delta$ respectively.
  \item \textit{Transition probability distribution} $P(\cdot|s_{t,c},a_{t,c})$: describes the state evolution given the current state and the executed action. It is composed by a deterministic part and a stochastic part: the downtilt varies deterministically as $\theta_{t+1,c}= \theta_{t,c} + a_{t,c}$, while the transition probabilities for $\textsc{Cov}_{t,c}, \textsc{Cap}_{t,c}, \textsc{Qual}_{t,c}$ are assumed to be stochastic and unknown.
  \item \textit{Reward} $r_{t,c} = \log(1 + \textsc{Cov}_{t,c}^2 +\textsc{Cap}_{t,c}^2 + \textsc{Qual}_{t,c}^2)$: consists of the squared log-sum of the coverage, capacity, and quality KPIs.
  \item \textit{Baseline dataset} $\mathcal{D}_{\pi_b} = \{(s_n,a_n,r_n,s'_{n})\}_{n=1}^N$: consists of a collection of state, action, reward, next state tuples, collected by a baseline policy $\pi_b$. More information and statistics on $\pi_b$ and $\mathcal{D}_{\pi_b}$ are presented in Section \ref{sec:experiments}.
  \item \textit{Agent policy} $\pi_w : \mathcal{S}\to \Delta_{\mathcal{A}}$: consists of a probabilitstic policy derived from the state-action value function $Q_w(s,a)$. 
   We assume that the RL agent executes action on a cell-by-cell basis, and the policy is trained independently across data coming from each cell $c = 1,\dots, \mathcal{C}$.
   The agent's policy selects actions by sampling from a policy distribution $\pi_w(\cdot|s) \sim \textrm{Cat}(\sigma\left(\mathbf{Q}_w(s)\right))$, where $\mathbf{Q}_w(s) = [Q_w(s,a_1),\dots,Q_w(s,a_K)]$, $\sigma(\cdot)$ denotes the softmax function, and $\textrm{Cat}(\cdot)$ denotes the categorical distribution.
\end{itemize}

\subsection{SPIBB-DQN for RET optimization}
\renewcommand{\thefootnote}{$\ddagger$} 
This section details the application of the SPIBB-DQN algorithm to the RET use case. The pseudocode of SPIBB-DQN for RET optimization is presented in Algorithm \ref{algo:SPIBB_RET}. 
\begin{algorithm}
\caption{SPIBB-DQN for RET optimization}
\begin{algorithmic}[1]
\vspace{6.5pt}\Algphase{Phase 1 - Data Collection}
\State \small{\textbf{Input:} Baseline policy $\pi_b$, time budget $T$.}
\State \textbf{Initialize:} Initial state $s_0 \sim p_0$.
    \For{$t = 0, \dots, T$}
        \For{$c = 1, \dots, \mathcal{C}$}
            \State Select $ a_{t,c} \sim \pi_b(\cdot|s_{t,c})$,
            \State Receive $r_{t,c} \sim R(\cdot|s_{t,c},a_{t,c})$, \State Transition to $s_{t+1,c} \sim P(\cdot|s_{t,c},a_{t,c})$,
            \State Append $\mathcal{D}_{\pi_b} \leftarrow \mathcal{D}_{\pi_b} \cup (s_{t,c},a_{t,c},r_{t,c},s'_{t,c})$.
        \EndFor
    \EndFor
\State \textbf{Return} $\mathcal{D}_{\pi_b}$.
\Algphase{Phase 2 - SPIBB-DQN Learning}
\State \textbf{Input:} Baseline dataset $\mathcal{D}_{\pi_b}$, and parameters $N_\wedge$, $N_B$, $N_{\text{epoch}}$.
\State \textbf{Initialize:} $Q$-function weights $w$ randomly.
\State Compute pseudo-counts $\widetilde{N}_{\mathcal{D}_{\pi_b}}(s,a)$ as in \eqref{eq:pseudo-count}.
\State Construct bootstrapped set $\mathfrak{B}_{\pi_{b}}$ as in \eqref{eq:bootstrapped}.
\For{$m = 1, \dots, N_{\text{epoch}}$}
     \For{$j = 1, \dots, \left \lfloor{N/N_B}\right \rfloor $}
     \State Sample batch $\{(s_{\beta},a_{\beta},r_{\beta},s'_{\beta})\}_{\beta = 1}^{N_B}$ from $\mathcal{D}_{\pi_b},$
        \State Set the target $y_{\beta}^{(j)}$ as in \eqref{eq:SPIBB_update}  for $\beta = 1,\dots, N_B$,
        \State \small{Update} $w \leftarrow w - \eta\sum_{\beta}^{N_B}\nabla_{w} \left(y_{\beta}^{(j)}-Q_{w}\left(s_{\beta}, a_{\beta}\right)\right)^{2}$
    \EndFor
\EndFor
\State \textbf{Return}: trained policy $\pi_w^\star$,  
\end{algorithmic}
\label{algo:SPIBB_RET}
\end{algorithm}
\\ The algorithm can be divided into two phases:
\begin{enumerate}
    \item \textbf{Data collection}: Collect $\mathcal{D}_{\pi_b}$ by letting the safe baseline policy $\pi_b$ interact with the environment $\mathcal{E}$. At each time step $t = 1,\dots,T$, and for each cell $c = 1, \dots,\mathcal{C}$, the tuple $(s_{t,c},a_{t,c},r_{t,c},s'_{t,c})$ is generated according to the RL interaction cycle and stored. This process is repeated for a total of $N = T\cdot\mathcal{C}$ dataset entries. 
    \item \textbf{SPIBB-DQN learning}: Learn a safe policy from the collected dataset $\mathcal{D}_{\pi_b}$. First, the state-action counts are computed. Since we deal with a continuous state space, we utilize the pseudo-count $\widetilde{N}_\mathcal{D}(s,a)$ for counting the state-action pairs in a dataset $\mathcal{D}$, defined as
    \begin{equation}
        \label{eq:pseudo-count}
        \widetilde{N}_\mathcal{D}(s,a) = \sum_{(s_i,a_i) \in \mathcal{D}} \lVert s-s_i \rVert_2 \mathbbm{1}\{a = a_i\},
    \end{equation}
    where $\left\Vert\cdot\right\Vert_2$ denotes the Euclidean norm. Based on the pseudo-count computation, the bootstrapped set $\mathfrak{B}_{\pi_{b}}$ is constructed. Subsequently, the learning proceeds in batches of length $N_B$ sampled from the dataset $\mathcal{D}_{\pi_b}$ for $N_{\text{epochs}}$. For each batch, the bootstrapped update on the $Q$-function is executed. At the end of the process, the trained policy $\pi_w^\star$ is returned.
\end{enumerate}

\section{Experimental setting and results}
\label{sec:experiments}
This section describes the experimental setup, the evaluation metrics, and our experimental results.
\subsection{Experimental setup}
\label{sec:experimental_setup}
We execute our experiments in a static simulated network environment. The network environment map is shown in Fig. \ref{fig:network_env}, and the simulation parameters are reported in Table \ref{tab:simulator_setup}. 
\begin{table}[ht!]
\centering
\caption{Simulator parameters.}
\begin{tabular}{lcl}
\hline
\textsc{Simulator parameter} & \textsc{Symbol} & \textsc{Value} \\ \hline 
\hline
Number of BSs & $\mathcal{B}$ & $7$ \\
Number of cells &$\mathcal{C}$& $21$ \\
Number of UE &$\mathcal{U}$ &$2000$ \\
Frequency &$f$  & $2$ GHz \\
Traffic volume average&$\mu_\tau$ & $20$ Mbps \\
Traffic volume variance &$\sigma^2_\tau$& $4$ Mbps \\
Antenna height&$h$ & $32$ m \\
Minimum downtilt angle&$\theta_{\text{min}}$ & $1^{\circ}$\\
Maximum downtilt angle &$\theta_{\text{max}}$& $16^{\circ}$ \\ \hline 
\label{tab:simulator_setup}
\end{tabular}
\end{table}
\\Two different baselines are used in our experiments: a \textit{RB baseline} $\pi_{_\text{RB}}$, and a \textit{ DQN baseline} $\pi_{_\text{DQN}}$. The RB baseline consists of a legacy domain-knowledge policy deployed in the real-world network, and is thus considered safe. The DQN baseline consists of a sub-optimal RL policy trained in the simulation environment for $500$ steps and stopped before convergence. In contrast to the RB baseline, the DQN baseline is stochastic and explores a larger portion of the state-action space. For both baseline policies, we conduct a set of experiments by applying Algorithm \ref{algo:SPIBB_RET}: we collect the datasets with size $|\mathcal{D}_{\pi_{\text{RB}}}| = |\mathcal{D}_{\pi_{\text{DQN}}}| = 500$, as in Phase $1$, and subsequently train the SPIBB-DQN model by constructing the bootstrapped sets $\mathfrak{B}_{\pi_{\text{RB}}}$, $\mathfrak{B}_{\pi_{\text{DQN}}}$, and running the bootstrapped $Q$-function updates, as in Phase $2$. We execute $K = 20$ independent runs for each experiment, reporting mean and standard deviation of the performance. We execute experiments for different values of $N_\wedge \in \{5, 10, 20, 30, 40, 50, 100, 150, 200, 300\}$ and offline dataset size $N \in \{25, 50, 100, 200, 300, 400, 500\}$. The experiments are executed according to the following method: for each episode, we reset the downtilt uniformly at random, i.e. at $t = k\cdot T_\textrm{episode}$, $k = 0,\dots,N_\textrm{episode}-1$, we set ${\theta_{t,c} \sim \textsc{U}\left(\theta_\textrm{min},\theta_\textrm{max}\right)}$, where $\textsc{U}(\cdot)$ is the discrete uniform distribution. We report the training and testing hyperparameters in Table \ref{tab:hyperparams}.
\begin{table}[htb!]
\centering
\caption{Training $\&$ Testing Hyperparameters.}
\begin{tabular}{lcl}
\hline
\textsc{Hyperparameter} &\textsc{Symbol} & \textsc{Value} \\ \hline 
\hline
Discount factor &$\gamma$ & $9\times 10^{-1}$ \\
Learning rate &$\eta$ & $1\times 10^{-3}$ \\
Batch size &$N_B$ & $50$ \\
Test episode length &$T_{\text{episode}}$ & $20$ \\
Number of evaluation episodes &$N_{\text{episode}}$& $25$ \\
Test epoch length &$T_{\text{epoch}}$ & $500$ \\
Number of training epochs &$N_{\text{epoch}}$& $20$ \\ \hline 
\label{tab:hyperparams}
\end{tabular}
\end{table}


Fig. \ref{figure:DQN} illustrates the ANN architecture used for the SPIBB policies. The ANN model takes as input a state-action pair $(s,a)$ and outputs the $Q$-value for $(s,a)$. 

It is composed of three hidden fully-connected layers, whose number of units is reported in parenthesis in Fig. \ref{figure:DQN}. The activation function for the hidden layers is the Rectified Linear Unit (ReLU), while for the output layer, we use a linear activation function.
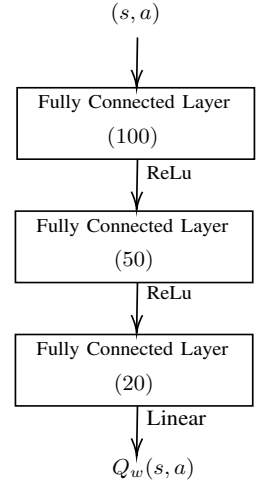
\begin{figure}
\centering
\tikzset{every picture/.style={line width=0.75pt}}
\resizebox{3.2cm}{6.4cm}{%
\begin{tikzpicture}[x=0.75pt,y=0.75pt,yscale=-1,xscale=1]
\draw    (80.5,28) -- (80.03,57) ;
\draw [shift={(80,59)}, rotate = 270.92] [color={rgb, 255:red, 0; green, 0; blue, 0 }  ][line width=0.75]    (10.93,-3.29) .. controls (6.95,-1.4) and (3.31,-0.3) .. (0,0) .. controls (3.31,0.3) and (6.95,1.4) .. (10.93,3.29)   ;
\draw    (80.5,179) -- (80.03,208) ;
\draw [shift={(80,210)}, rotate = 270.92] [color={rgb, 255:red, 0; green, 0; blue, 0 }  ][line width=0.75]    (10.93,-3.29) .. controls (6.95,-1.4) and (3.31,-0.3) .. (0,0) .. controls (3.31,0.3) and (6.95,1.4) .. (10.93,3.29)   ;
\draw   (9.5,59) -- (149.5,59) -- (149.5,103) -- (9.5,103) -- cycle ;
\draw    (80.5,103) -- (80.03,132) ;
\draw [shift={(80,134)}, rotate = 270.92] [color={rgb, 255:red, 0; green, 0; blue, 0 }  ][line width=0.75]    (10.93,-3.29) .. controls (6.95,-1.4) and (3.31,-0.3) .. (0,0) .. controls (3.31,0.3) and (6.95,1.4) .. (10.93,3.29)   ;
\draw    (80.5,255) -- (80.03,284) ;
\draw [shift={(80,286)}, rotate = 270.92] [color={rgb, 255:red, 0; green, 0; blue, 0 }  ][line width=0.75]    (10.93,-3.29) .. controls (6.95,-1.4) and (3.31,-0.3) .. (0,0) .. controls (3.31,0.3) and (6.95,1.4) .. (10.93,3.29)   ;
\draw   (8.5,135) -- (148.5,135) -- (148.5,179) -- (8.5,179) -- cycle ;
\draw   (8.5,211) -- (148.5,211) -- (148.5,255) -- (8.5,255) -- cycle ;
\draw (64,285.4) node [anchor=north west][inner sep=0.75pt]    {$Q_{w}( s,a)$};
\draw (63,5.4) node [anchor=north west][inner sep=0.75pt]    {$( s,a)$};
\draw (84.5,180) node [anchor=north west][inner sep=0.75pt]   [align=left] {{\small ReLu}};
\draw (20.5,62) node [anchor=north west][inner sep=0.75pt]   [align=left] {{\small Fully Connected Layer}\\};
\draw (61,81.4) node [anchor=north west][inner sep=0.75pt]    {$( 100)$};
\draw (84.5,106) node [anchor=north west][inner sep=0.75pt]   [align=left] {{\small ReLu}};
\draw (84.5,256) node [anchor=north west][inner sep=0.75pt]   [align=left] {Linear};
\draw (19.5,138) node [anchor=north west][inner sep=0.75pt]   [align=left] {{\small Fully Connected Layer}\\};
\draw (65,157.4) node [anchor=north west][inner sep=0.75pt]    {$( 50)$};
\draw (19.5,214) node [anchor=north west][inner sep=0.75pt]   [align=left] {{\small Fully Connected Layer}\\};
\draw (65,233.4) node [anchor=north west][inner sep=0.75pt]    {$( 20)$};
\end{tikzpicture}}
\caption{SPIBB network architecture.}
    \label{figure:DQN}
\end{figure}

\subsection{Evaluation metrics}
\begin{figure*}[ht!]
  \centering
  \includegraphics[width=0.975\linewidth]{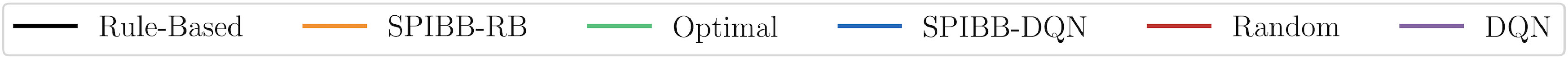} \par
  \subfloat[Average Network Reward vs $t$][Average Network Reward vs $t$.]{\includegraphics[width=.331\textwidth]{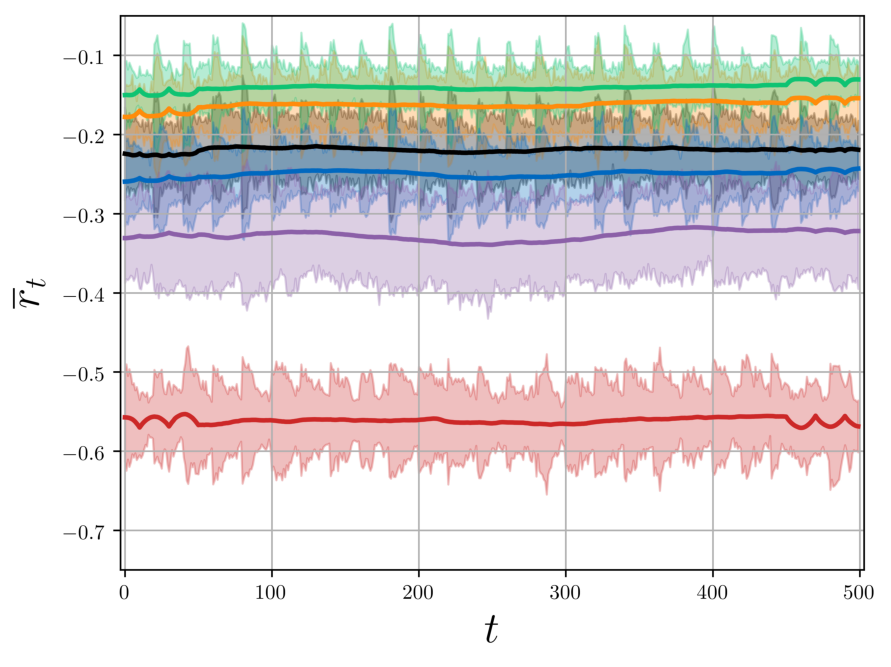}\label{fig:exp_a}} 
  \subfloat[][CVaR Average Network Reward vs $t$.]{\includegraphics[width=.331\textwidth]{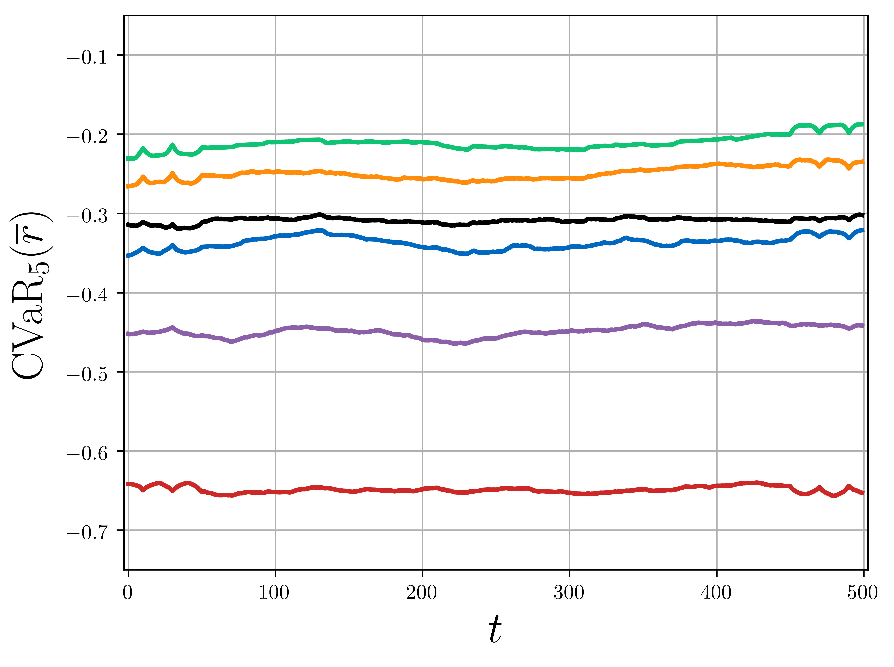}\label{fig:exp_b}}
  \subfloat[Minimum cell reward][ Minimum Cell Reward vs $t$.]{\includegraphics[width=.331\textwidth]{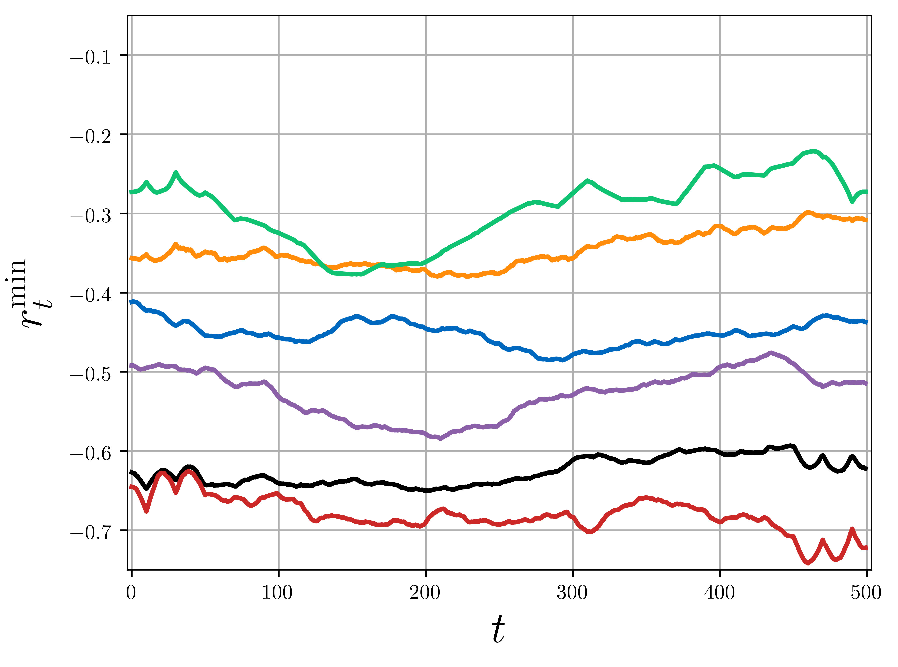}\label{fig:exp_c}}\par
  \subfloat[][ Average Network Reward vs $N$.]{\includegraphics[width=.331\textwidth]{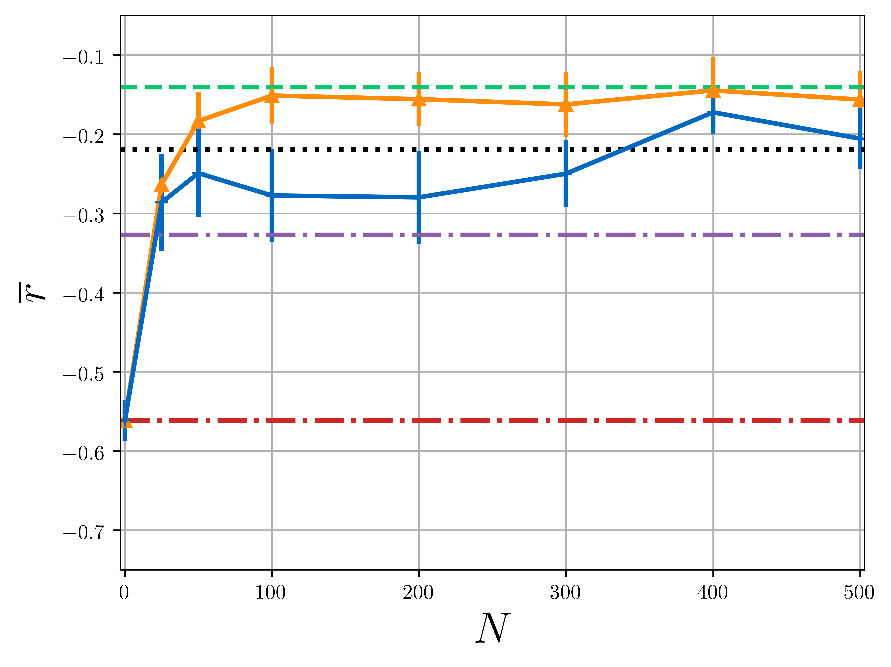}\label{fig:exp_d}}
  \subfloat[][CVaR Average Network Reward vs $N$.]{\includegraphics[width=.331\textwidth]{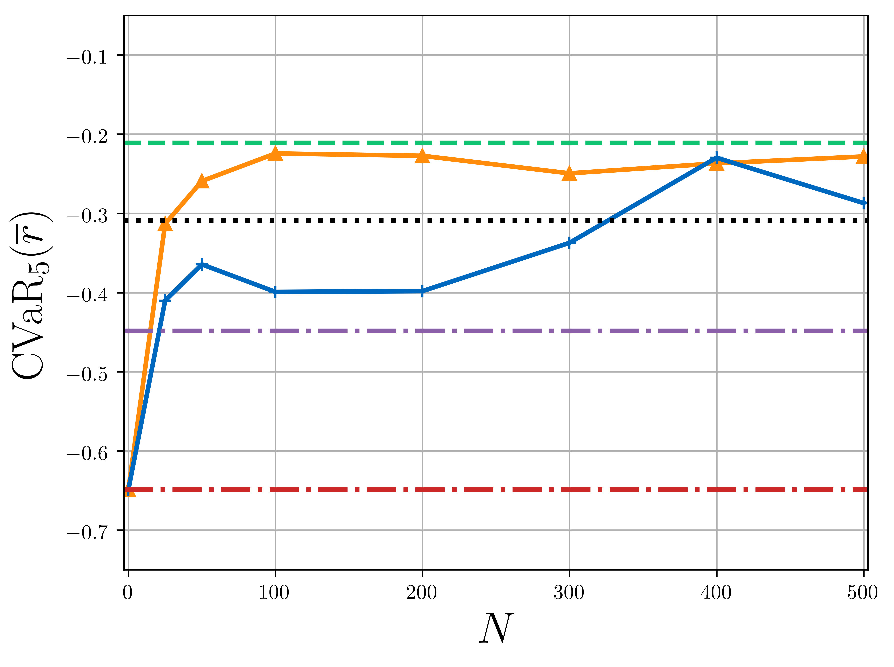}\label{fig:exp_e}}
  \subfloat[][ Minimum Cell Reward vs $N$.]{\includegraphics[width=.331\textwidth]{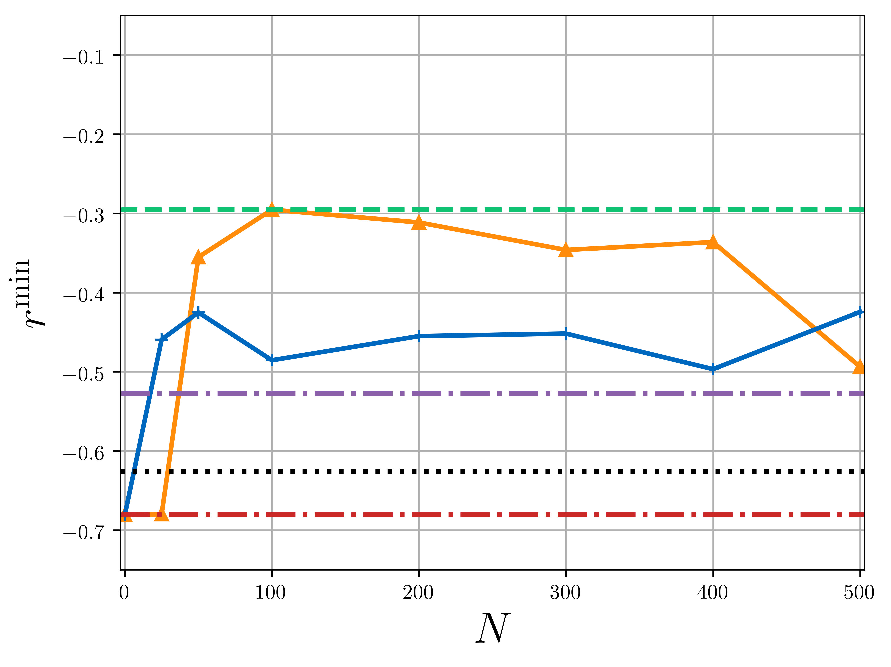}\label{fig:exp_f}}\par
  \subfloat[][Average Network Reward vs $N_\wedge$.]{\includegraphics[width=.331\textwidth]{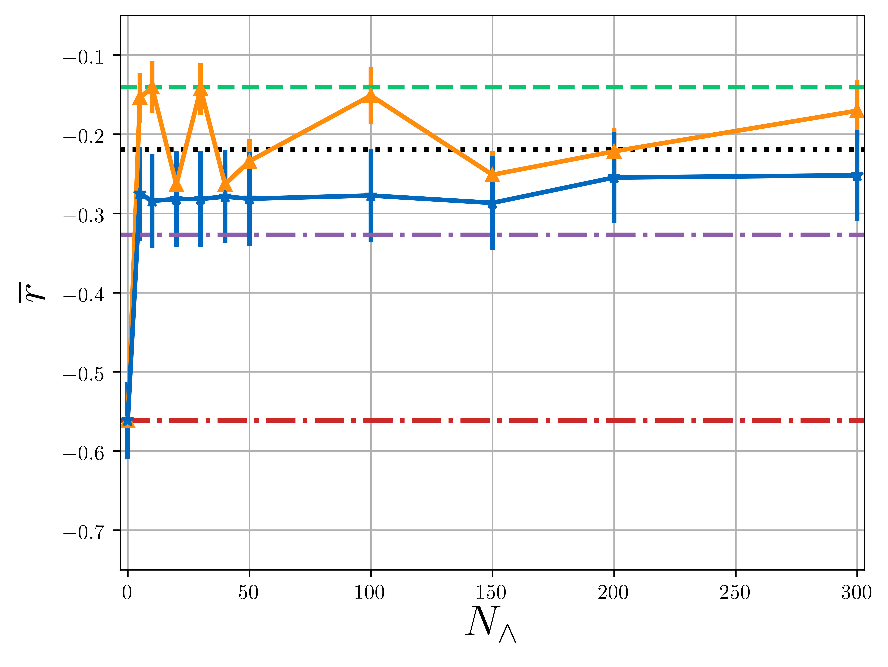}\label{fig:exp_g}}
  \subfloat[][CVaR Average Network Reward vs $N_\wedge$.]{\includegraphics[width=.331\textwidth]{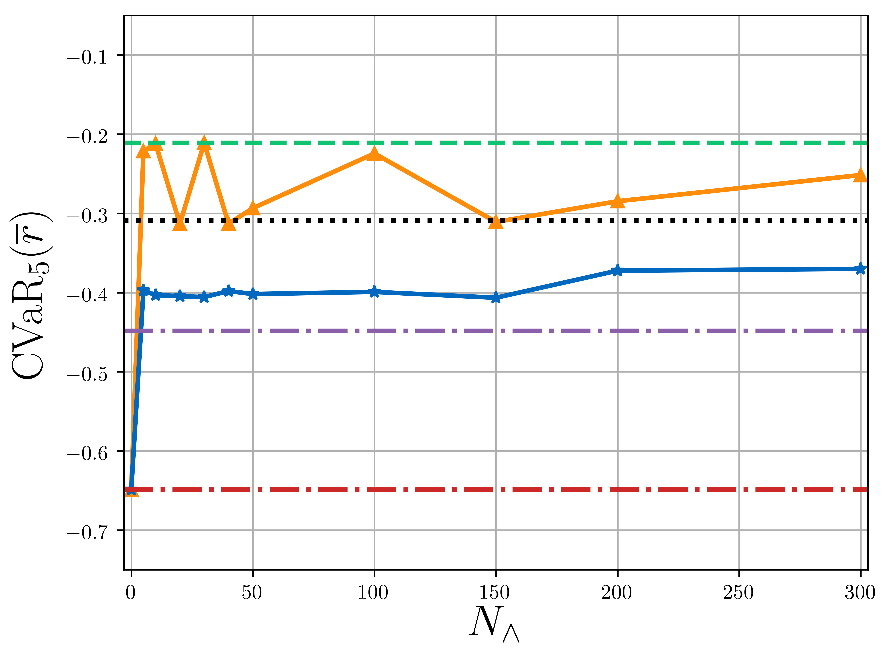}\label{fig:exp_h}}
  \subfloat[][Minimum Cell Reward vs $N_\wedge$.]{\includegraphics[width=.331\textwidth]{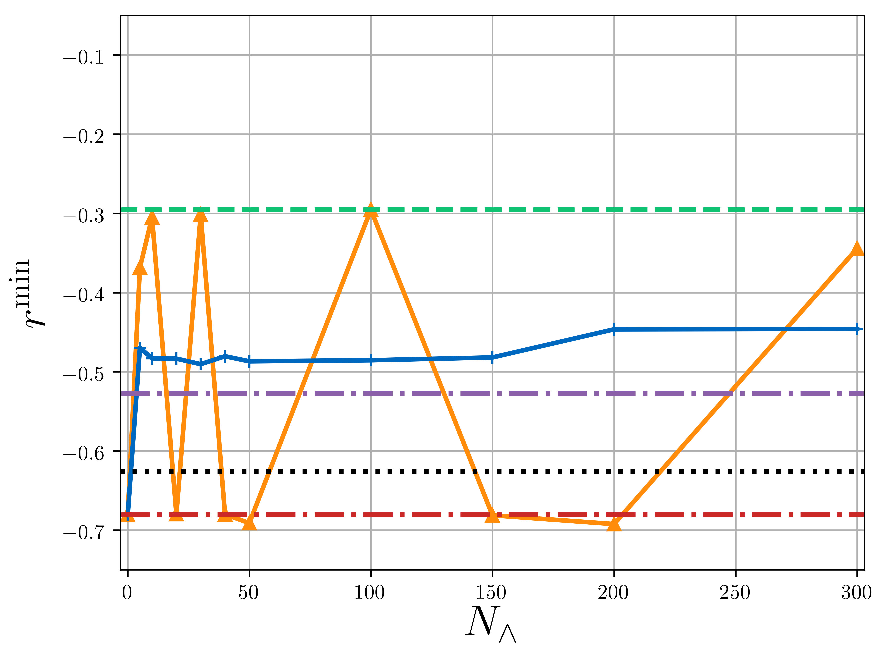}\label{fig:exp_i}}\hfill
  \caption{Experimental results.}\label{fig:experiments_tilt}
\end{figure*}

For evaluation, we let the policies interact with the simulated network environment on unseen test environment configurations. The policies considered for the evaluations are: \textit{RB} and \textit{DQN} baseline policies, the \textit{SPIBB-RB} and \textit{SPIBB-DQN} policies, trained on the RB and DQN dataset respectively, the \textit{uniform random} policy, and the (empirical) \textit{optimal} policy. We utilize the following evaluation metrics:
\begin{enumerate}
    \item \textit{Average network reward:} $\bar{r}_t$, a general performance indicator considering the average over the cells' performance of the controlled network area, defined as:
\begin{equation*}\label{eq:mean_reward}
\bar{r}_t \triangleq \frac{1}{\mathcal{C}} \sum_{c = 1}^\mathcal{C} r_{t,c}.
\end{equation*}
With a slight abuse of notation we denote by ${\bar{r} = \frac{1}{T_\textrm{epoch}} \sum_{t = 1}^{T_\textrm{epoch}} \bar{r}_t}$ the time average of $\bar{r}_t$.
\item \textit{Conditional Value-at-Risk:} $\textrm{CVaR}_a(r)$, the average reward obtained from the $a\%$ worst runs, defined as 
\begin{equation*}\label{eq:CVaR}
\textrm{CVaR}_a(r) \triangleq \mathbb{E}[r\mid r \leq \textrm{VaR}_a(r)],
\end{equation*}
where VaR denotes the Value at Risk, defined as $\textrm{VaR}_a(r) \triangleq \min\{\rho \in\mathbb{R} : \mathbb{P}[r\leq \rho]\geq a\}$.
\item \textit{Minimum cell reward:} $r^{\text{min}}_{t}$, considering the reward performance of the worst cell, defined as
\begin{equation*}\label{eq:min_reward}
r^{\text{min}}_{t} \triangleq \min_{c\in \mathcal{C}} r_{t,c}.
\end{equation*}
Finally, we denote by $\bar{r}^{\text{min}} =\frac{1}{T_\textrm{epoch}} \sum_{t = 1}^{T_\textrm{epoch}} r^{\text{min}}_{t}$ the time average of $r_{t,c}$.
\end{enumerate}

\subsection{Results and discussion}
\label{sec:results_and_discussion}
The results of our experiments are shown in Fig. \ref{fig:experiments_tilt}. The columns contain the result for the average network reward, the $5 \%$ CvaR of the average network performance, and the minimum cell reward. The first row shows an example of performance evaluation for fixed data size $N = 300$ and safety parameter $N_\wedge = 100$. The second row shows the performance of the considered policies when varying the data size $N$, and for fixed safety parameter $N_\wedge = 100$. The third row shows the performance of the considered policies when varying the safety parameter $N_\wedge$ and for fixed data size $N = 100$. Finally, in Table \ref{tab:results} we report the experimental results for the (first column contains results in terms of mean and standard deviation) across the $K = 20$ runs for $N = N_\wedge = 100$. 

\begin{table}[ht!]
\fontsize{9.3}{9.3}\selectfont
\centering
\caption{Experimental Results for $N = N_\wedge = 100$}
\begin{tabular}{cccc}\hline
\textbf{Policy} & $\bar{r}$& $\textrm{CVaR}_5(\bar{r}) $ & $\bar{r}_\textrm{min}$ \\ \hline \hline
\textbf{Random}  & $-0.562 \pm 0.049$ & $-0.649 $ & $-0.681$ \\
\textbf{DQN} & $-0.327\pm0.061$ & $-0.448$ & $-0.527$ \\
\textbf{SPIBB-DQN}  & $-0.278\pm0.059$ & $-0.399$ & $-0.486$ \\
\textbf{Rule-Based} & $-0.219\pm0.046$ & $-0.309$ & $-0.626$ \\
\textbf{SPIBB-RB} & $-0.151\pm0.036$  & $-0.224$ &  $0.295$  \\
\textbf{Optimal} &  $-0.141\pm0.033$ & $-0.211$ & $-0.295$ \\ \hline 
\label{tab:results}
\end{tabular}
\end{table}
In the following points, we summarize and discuss the key conclusions from the experimental results:
\begin{enumerate}
    \item \textit{The SPIBB policies outperform the respective baselines:} from Figs. \ref{fig:exp_a}, \ref{fig:exp_b}, \ref{fig:exp_c}, we may observe that safe improvement over the baselines is achieved at any time step of the evaluation. Also, Figs. \ref{fig:exp_d}, \ref{fig:exp_e}, \ref{fig:exp_f} show that the SPIBB policies outperform the respective baseline in terms of $\bar{r}_t$, $\textrm{CVaR}_5(r)$, and $\bar{r}^{\text{min}}$. The first metric represents the overall network performance, while the second and third metrics embody worst-case scenarios. The result validates the effectiveness of the SPIBB, providing safe policy improvement over the respective baselines for average and different worst-case performance. At the best hyperparameter configuration ($N = 100$, $N_\wedge = 10$), the SPIBB RB achieves a normalized improvement over the RB baseline of $87.2\%$ (normalization is taken with respect to the empirical optimal reward). 
    \item \textit{The SPIBB policies performance increases with the dataset size:} a larger amount of data allows learning better policies during the offline learning process. This is confirmed for both the RB policy, which is closer to a deterministic policy, and the DQN policy, that exhibits a higher degree of randomization. Amazingly, we observe that even small data sizes ($N\geq 50$) allow the SPIBB policies to outperform the respective baselines. This highlights the high sample efficiency of the SPIBB method and its robustness to different types of baselines. 
    \item \textit{The SPIBB-RB policy outperforms the SPIBB-DQN policy:} this is a direct consequence of three facts: first, the RB baseline policy outperforms the DQN baseline policy. The DQN baseline is stopped before convergence and thus has worse performance than the RB baseline, as confirmed by the majority of the results in Fig. \ref{fig:experiments_tilt}. Second, the RB policy collects high-performance trajectories that are stored in the RB baseline dataset and used in the offline training of the SPIBB-RB policy. Third, the RB action probabilities, used to bootstrap the state-action pairs on the RB bootstrapped set, are closer to the ones of the optimal policy when compared to the DQN policy.
    \item \textit{The SPIBB policy is highly sensitive to the safety parameter $N_\wedge$:} from Figs. \ref{fig:exp_g}, \ref{fig:exp_h}, \ref{fig:exp_i}, we can observe how different values of the safety parameter $N_\wedge$ significantly affect the SPIBB performance. Generally, we expect that higher values of the safety parameter make the policy safer but more prone to sub-optimal performance. On the one hand, we observe how the SPIBB-DQN policy exhibits a more stable behavior to variations of this parameter: it outperforms the DQN baseline for all experimented $N_\wedge$. This may be due to the high degree of randomization of the DQN baseline and the subsequent increased diversity of the dataset. On the other hand, the SPIBB-RB baseline shows a very unstable behavior, especially for the minimum cell reward performance. This may be due to the fact that the RB baseline is mostly deterministic and a very biased composition of the bootstrapped set could make the performance highly sensitive to $N_\wedge$. 
\end{enumerate}

\balance
\section{Conclusion}
\label{sec:conclusions}
In this paper, we devise and evaluate a safe downtilt update policy using SPIBB, a recent and effective SRL method based on the use of a safe baseline policy and batches of data containing trajectories collected by the baseline policy. We demonstrate the validity of such SRL approach through extensive experimental evaluation on a simulated mobile network environment for different baseline policies and dataset sizes. We show that SPIBB is able to produce a policy that outperforms different baseline policies and achieve safety at any instant of the execution, unlike standard RL methods. Possible future works include exploring the use of SRL methodologies for the optimization of multi-antenna systems (e.g., 5G MIMO systems), experimentation in more complex scenarios and realistic scenarios (e.g. testing scalability by increasing the network size and UEs), or using different metrics (e.g. throughput, packet loss, SINR, etc.) to model the network environment and the reward function. 

\section*{Acknowledgement}
\label{sec:ack}
This work was partially supported by the Wallenberg AI, Autonomous Systems and Software Program (WASP) funded by the Knut and Alice Wallenberg Foundation.
\bibliography{biblio.bib}

\begin{thebibliography}{10}
\providecommand{\url}[1]{#1}
\csname url@samestyle\endcsname
\providecommand{\newblock}{\relax}
\providecommand{\bibinfo}[2]{#2}
\providecommand{\BIBentrySTDinterwordspacing}{\spaceskip=0pt\relax}
\providecommand{\BIBentryALTinterwordstretchfactor}{4}
\providecommand{\BIBentryALTinterwordspacing}{\spaceskip=\fontdimen2\font plus
\BIBentryALTinterwordstretchfactor\fontdimen3\font minus
  \fontdimen4\font\relax}
\providecommand{\BIBforeignlanguage}[2]{{%
\expandafter\ifx\csname l@#1\endcsname\relax
\typeout{** WARNING: IEEEtran.bst: No hyphenation pattern has been}%
\typeout{** loaded for the language `#1'. Using the pattern for}%
\typeout{** the default language instead.}%
\else
\language=\csname l@#1\endcsname
\fi
#2}}
\providecommand{\BIBdecl}{\relax}
\BIBdecl

\bibitem{sutton2018reinforcement}
R.~S. Sutton and A.~G. Barto, \emph{Reinforcement learning: An
  introduction}.\hskip 1em plus 0.5em minus 0.4em\relax MIT press, 2018.

\bibitem{Balevi2019OnlineLearning}
E.~Balevi and J.~G. Andrews, ``{Online Antenna Tuning in Heterogeneous Cellular
  Networks with Deep Reinforcement Learning},'' \emph{IEEE Transactions on
  Cognitive Communications and Networking}, 2019.

\bibitem{Guo2013Spectral-andLearning}
W.~Guo, S.~Wang \emph{et~al.}, ``{Spectral-and energy-efficient antenna tilting
  in a HetNet using reinforcement learning},'' \emph{IEEE Wireless
  Communications and Networking Conference, WCNC}, 2013.

\bibitem{Razavi10}
R.~Razavi, S.~Klein, and H.~Claussen, ``{A fuzzy reinforcement learning
  approach for self-optimization of coverage in {LTE} networks},'' \emph{Bell
  Labs Technical Journal}, 2010.

\bibitem{Fan2014Self-optimizationLearning}
S.~Fan, H.~Tian, and C.~Sengul, ``{Self-optimization of coverage and capacity
  based on a fuzzy neural network with cooperative reinforcement learning},''
  \emph{Eurasip Journal on Wireless Communications and Networking}, 2014.

\bibitem{Tall15}
A.~Tall, Z.~Altman, and E.~Altman, ``{Self-optimizing Strategies for Dynamic
  Vertical Sectorization in LTE Networks},'' in \emph{{IEEE Wireless
  Communications and Networking Conference, WCNC}}, 2015.

\bibitem{Buenostado17}
V.~{Buenestado}, M.~{Toril} \emph{et~al.}, ``Self-tuning of remote electrical
  tilts based on call traces for coverage and capacity optimization in {LTE},''
  \emph{IEEE Transactions on Vehicular Technology}, 2017.

\bibitem{Garcia2015}
J.~Garc{\'{i}}a and F.~Fern{\'{a}}ndez, ``{A comprehensive survey on safe
  reinforcement learning},'' \emph{Journal of Machine Learning Research}, 2015.

\bibitem{Lange2012BatchLearning}
S.~Lange, T.~Gabel, and M.~Riedmiller, ``{Batch reinforcement learning},'' in
  \emph{Adaptation, Learning, and Optimization}.\hskip 1em plus 0.5em minus
  0.4em\relax Springer Verlag, 2012.

\bibitem{Thomas2015HighImprovement}
P.~S. Thomas, G.~Theocharous, and M.~Ghavamzadeh, ``{High confidence policy
  improvement},'' \emph{32nd International Conference on Machine Learning,
  ICML}, 2015.

\bibitem{Laroche2019SafeBootstrapping}
R.~Laroche, P.~Trichelair, and R.~T. Des~Combes, ``{Safe policy improvement
  with baseline bootstrapping},'' \emph{36th International Conference on
  Machine Learning, ICML}, 2019.

\bibitem{Vannella20}
F.~Vannella, J.~Jeong, and A.~Prouti{{e}}re, ``{O}ff-policy {L}earning for
  {R}emote {E}lectrical {T}ilt {O}ptimization,'' \emph{IEEE 92nd Vehicular
  Technology Conference, VTC Fall}, 2020.

\bibitem{Vairamuth20}
V.~Raghunath, ``Antenna tilt optimization using {R}einforcement {L}earning with
  doubly-safe exploration,'' {MS}c Thesis, KTH Royal Institute of Technology,
  2020.

\bibitem{Watkins89}
C.~J. C.~H. Watkins, ``Learning from delayed rewards,'' Ph.D. dissertation,
  King's College, Cambridge, UK, May 1989.

\bibitem{Mnih15}
V.~Mnih, K.~Kavukcuoglu \emph{et~al.}, ``Human-level control through deep
  reinforcement learning,'' \emph{Nature}, 2015.

\end{thebibliography}

\end{document}